%% file: CAP-CoT.tex
\documentclass[a4paper,fleqn]{cas-dc}

\usepackage[numbers]{natbib}
\usepackage{amsmath}
\usepackage{amssymb}
\usepackage{amsfonts}
\usepackage{algorithmic}
\usepackage{algorithm}
\usepackage{array}
\usepackage[caption=false, font=normalsize, labelfont=sf, textfont=sf]{subfig}
\usepackage{textcomp}
\usepackage{stfloats}
\usepackage{url}
\usepackage{verbatim}
\usepackage{graphicx}
\usepackage{booktabs}
\usepackage{multirow}
\usepackage{makecell}
\usepackage{diagbox}
\usepackage{tcolorbox}
\usepackage{longtable}
\usepackage{xcolor}
\usepackage{colortbl}
\usepackage{acro}
\usepackage{balance}
\input{CQILABInclusion/CQILAB-Notations}

\input{CQILABInclusion/CQILAB-Acronyms}

\input{LatexInclusion/CAP-CoT-Acronyms}
\input{LatexInclusion/CAP-CoT-Notations}

\def\tsc#1{\csdef{#1}{\textsc{\lowercase{#1}}\xspace}}
\tsc{WGM}
\tsc{QE}

\begin{document}
	\let\WriteBookmarks\relax
	\def\floatpagepagefraction{1}
	\def\textpagefraction{.001}
	\shorttitle{CAP-CoT}
	\shortauthors{Chen et~al.}
	
	\title [mode = title]{CAP-CoT: Cycle Adversarial Prompt for Improving Chain of Thoughts in LLM Reasoning}                      
	
	
	\author[1]{Shuxu Chen}
	
	\author[2]{Yitian Zhou}

	\author[2]{Jiaquan Zhang}

	\author[2]{Haoyu Bian}

    \author[3]{Wenrui Hu}
    
	\author[4]{Aming Wu}

	\author[1]{Sungyoung Lee}

	\author[2]{Chaoning Zhang}
	\cormark[1]
	\ead{chaoningzhang1990@gmail.com}
	
	\author[1]{Hyundong Shin}
	\cormark[1]
	\ead{hshin@khu.ac.kr}
	
	\affiliation[1]{organization={Department of Electronics and Information Convergence Engineering, Kyung Hee University},
		country={Republic of Korea}}
	
	\affiliation[2]{organization={University of Electronic Science and Technology of China},
		city={Chengdu},
		country={China}}
		
	\affiliation[4]{organization={East China Jiaotong University},
		city={Nanchang},
		country={China}}
	
	\affiliation[4]{organization={Hefei University of Technology},
		city={Hefei},
		country={China}}
	
	\cortext[1]{Corresponding authors}

	\begin{abstract}
		Chain-of-Thought (CoT) prompting has emerged as a simple and effective way to elicit step-by-step solutions from large language models (LLMs). However, CoT reasoning can be unstable across runs on long, multi-step problems, leading to inconsistent answers for unchanged task. Most prior work focuses on improving the forward reasoning chain within a single pass, with less attention to iterative and contrastive correction.
		To address this gap, we propose CAP-CoT, a Cycle Adversarial Prompt optimization framework designed to improve both CoT reasoning accuracy and stability of a single deployed solver. In each cycle, a forward solver generates candidate reasoning chains, an adversarial challenger constructs plausible but deliberately flawed chains using targeted error strategies, and a feedback agent contrasts the two chains and produces step-aligned structured feedback. This feedback closes the optimization loop in two directions, including updating the solver prompt based on errors exposed by the challenger, and updating the challenger prompt to generate increasingly targeted errors in subsequent cycles.
		Unlike safety-oriented adversarial prompting such as jailbreak or prompt-injection attacks, our adversarial component is task-semantic and aims to expose logical vulnerabilities in reasoning chains.
		Experiments across six benchmarks and four LLM backbones demonstrate that within two to three adversarial prompt optimization cycles, CAP-CoT consistently reduces variability across runs while improving reasoning accuracy and robustness to prompt perturbations.
	\end{abstract}
	

	
	\maketitle
	\acresetall	
	\section{Introduction}
	
	\Acp{LLM} have shown strong performance across a wide range of reasoning tasks, particularly with prompt-based methods such as \ac{CoT} prompting, which encourages step-by-step solutions by structuring intermediate reasoning before arriving at a final answer~\cite{WWSDBCZ:22:NEURIPS,KGSRMYI:22:NEURIPS}. Despite this progress, higher \ac{CoT} accuracy does not always translate into robust reasoning behavior. As problems grow longer and more complex, performance degrades substantially even for strong models~\cite{BYLRKAS:25:ICML}, revealing that the apparent capability measured on standard benchmarks may not generalize to harder or more varied instances. Moreover, \ac{CoT}-based approaches remain sensitive to prompt-level variation and contextual perturbations~\cite{YCZWHLEZ:25:ARXIV}. For instance, reordering premises without changing the underlying task can lead to large performance drops~\cite{XCRXWDZ:24:ICML}, and irrelevant context can distract the model's reasoning and reduce arithmetic accuracy~\cite{MYEHLSWL:25:EMNLP}. This sensitivity makes it difficult to distinguish which reasoning steps are truly robust from those that are only accidentally correct, especially on multi-step problems where early deviations propagate and compound throughout the chain. 
	
	Most prior work addresses these limitations by improving the \emph{forward} reasoning trace, for example, by refining prompts, sampling multiple trajectories, or searching over structured reasoning spaces~\cite{XWSQVLEN:23:ICLR,YDZGTGCN:23:NEURIPS,BMBKRGMP:24:AAAI,TYSZLW:26:NeurIPS,ZZCWHZ:26:ARXIV,ZSZWZHZPSBL:26:ARXIV}. While effective, these methods primarily optimize what the model should do next along a single forward direction and generally lack a systematic mechanism for exposing latent failure modes and converting them into targeted corrections. 
	A complementary perspective is that \ac{CoT} demonstrations may not reliably teach models how to reason~\cite{BSMWYLSJ:23:ACL}. When a model performs multi-step reasoning, errors within an intermediate step can derail subsequent steps and undermine final accuracy and reliability, which motivates explicit mechanisms to detect and repair such fragile reasoning~\cite{LFZLMRH:23:NEURIPS,TMEPB:23:NEURIPS}. Recent studies on efficient representation and retrieval also suggest that improving intermediate structure alone is insufficient without robustness-aware modeling~\cite{JGCQWXJO:25:ACL,JOSJZWQL:25:ACL}.
	
	A natural direction emerging from these observations is to learn from comparisons rather than exclusively from correct solutions, that is, to exploit the signal contained in both good and bad reasoning chains simultaneously. This intuition aligns with preference-based learning, in which human comparisons of better and worse outputs are used to train a reward model that drives subsequent improvement~\cite{OWJWAPRC:22:NEURIPS}. A related principle appears in representation learning, where supervised contrastive learning improves generalization by pulling positive examples together while repelling negative ones~\cite{KTWAPILC:20:NEURIPS}. Together, these ideas suggest that explicit contrast between correct and incorrect candidates can provide a strong learning signal beyond purely positive demonstrations, motivating contrastive prompting methods that show the model what to avoid alongside what to do in the prompt~\cite{CGTPB:23:ARXIV}.
	
	While static contrasts of this kind are effective, they can be limited because the negative examples are fixed and may become less informative as the reasoning model improves, particularly in long and complex chains where the space of meaningful errors shifts as earlier weaknesses are resolved. Thus, an iterative \emph{cycle} is proposed to address this, which continually generates targeted hard negatives calibrated to the model's current behavior while converting newly exposed weaknesses into explicit and actionable prompt updates. Iterative refinement of this form has proven broadly effective for improving \ac{LLM} behaviors at the prompt level~\cite{MSTGHMPC:23:NEURIPS,SCGNKY:23:NEURIPS,YWLLCZ:24:ICLR,JZJYZFTC:25:ICLR,ZZCHZLM:26:ARXIV}, reinforcing the case that closes the loop between error generation and prompt correction.
	
	Some research also improves prompts by using feedback signals to iteratively search for better instructions. OPRO~\cite{YWLLCZ:24:ICLR} treats the prompt as a text to be rewritten by the \ac{LLM} itself guided by past performance. \ac{PAgent}~\cite{XWCZWFBH:24:ICLR} frames prompt optimization as a planning and search problem guided by error feedback from task examples, reporting substantial gains over gradient-free baselines. These methods demonstrate that iterative, error-aware prompt updates can be highly effective. However, they rely on scalar correctness signals or externally specified error categories to drive updates; the feedback is not contrastively grounded in plausible erroneous chains that specifically target the model's current reasoning weaknesses. As a result, the optimization signal does not directly encode the structure of the mistakes a solver is prone to making.
	
	Inspired by these learning patterns, we propose \ac{CAP-CoT}, an optimization framework that strengthens \ac{CoT} reasoning through repeated contrast between a solver and an evolving challenger. \ac{CAP-CoT} consists of three role agents: a \emph{solver} that produces a forward reasoning chain, an \emph{adversarial challenger} that constructs plausible but flawed chains using controlled error strategies, and a \emph{feedback} agent that contrasts the two chains and outputs step-aligned, actionable instructions. Crucially, the feedback closes the cycle in both directions. It not only updates the solver prompt to repair fragile steps exposed by the challenger, but also updates the challenger prompt based on the solver's current behavior so that subsequent cycles generate increasingly targeted hard errors. We emphasize that the term \emph{adversarial} here is task-semantic, referring to the construction of deceptive but plausible reasoning errors, which is distinct from adversarial prompting in \ac{LLM} safety that typically denotes jailbreak prompts or prompt-injection attacks aimed at eliciting prohibited behaviors \cite{EPSFCRL:22:EMNLP,MMLPYXZW:24:ICML}. 
	By making the contrast evolve cycle by cycle in response to the model's current reasoning, \ac{CAP-CoT} systematically discovers and repairs reasoning weaknesses that static demonstrations may fail to surface, improving both accuracy and robustness.
	Our contributions are summarized as follows:
	\begin{itemize}
		\item We propose \ac{CAP-CoT}, a cycle-based adversarial prompt optimization framework that strengthens \ac{CoT} reasoning through adaptive contrast between correct and erroneous chains and treats prompt improvement as a closed loop between a solver and an adversarial challenger. The challenger starts from a lightweight error taxonomy and evolves over cycles to generate targeted hard negative reasoning chains for adversarial optimization.
		
		\item We introduce a feedback agent and form bidirectional \ac{SFPR} step in which the feedback agent simultaneously produces step-aligned improvement directives for the solver and targeted attack directives for the challenger, making the optimization loop transparent and interpretable.
		
		\item We conduct extensive evaluations of \ac{CAP-CoT} on multiple reasoning benchmarks under different \ac{LLM} backbones, with variance analysis provided in ablation experiments to characterize stability across runs, demonstrating improvements in accuracy and robustness.
	\end{itemize}
	
	The rest of this paper is organized as follows. Section~\ref{sec:2} reviews related work. Section~\ref{sec:3} presents the proposed \ac{CAP-CoT} framework. Section~\ref{sec:4} reports experimental results and analysis. Section~\ref{sec:5} concludes the paper.
	
	\section{Related Work}
	\label{sec:2}
	Our work builds on three main lines of research: \ac{CoT} reasoning, contrastive and adversarial prompting for robustness and stability, and feedback-driven prompt optimization.
	
	\subsection{Chain-of-Thought Reasoning}
	
	As \acp{LLM} have scaled in both capacity and capability, they demonstrate increasingly strong performance on complex reasoning tasks with suitable prompting strategies \citep{WZZWFHC:25:ARXIV,ZYLCFH:26:ARXIV}. Reasoning chain prompting improves \ac{LLM} reasoning by encouraging explicit, step-by-step solutions and has become a standard prompting primitive for arithmetic and symbolic tasks~\citep{WWSDBCZ:22:NEURIPS,KGSRMYI:22:NEURIPS}. Beyond single-chain \ac{CoT}, many methods strengthen test-time reasoning by aggregating multiple candidate traces or expanding the search space with structured exploration~\citep{ZSZWZHZPSBL:26:ARXIV}. 
	Self-Consistency samples diverse \ac{CoT} traces and selects the most consistent answer, often outperforming greedy decoding with modest changes to inference~\citep{XWSQVLEN:23:ICLR}. \Ac{ToT} treats intermediate reasoning steps as search states and performs deliberate exploration with evaluation and backtracking~\citep{YDZGTGCN:23:NEURIPS}. \Ac{GoT} generalizes linear or tree-structured traces to a graph of intermediate thoughts, enabling flexible composition and refinement through dependency edges~\citep{BMBKRGMP:24:AAAI}. More recently, \Ac{FoT} further scales test-time compute by integrating multiple reasoning trees and using consensus-style selection to revisit and correct flawed branches~\citep{BHLTW:25:ICML}. \Ac{AoT} decomposes complex problems into self-contained atomic subquestions with a Markov-style progression, reducing long-horizon interference and serving as a plug-in for test-time scaling~\citep{TYSZLW:26:NeurIPS}.
	In parallel, long-context benchmarks such as LongBench highlight that reasoning quality can degrade when models must integrate evidence over very long inputs, motivating methods that are robust to noisy or lengthy contexts~\citep{BZLJTHZD:24:ACL}.
	In multimodal settings, recent work further extends \ac{CoT} to visual reasoning, emphasizing alignment between modalities and intermediate reasoning faithfulness~\citep{ZZMJZZZSCZ:26:ARXIV,WCSZHZLMZQYY:26:ARXIV}

	\subsection{Contrastive and Adversarial Prompting}
	
	Adversarial training has long been employed to improve model robustness and generalization by constructing challenging counterexamples and encouraging models to handle harder cases. 
	Existing research has included models across modalities, including CycleGAN \citep{ZPIEB:17:ICCV} with unpaired translation, StyleGAN3 \citep{TKAMSLHJ:21:NEURIPS} with frequency-domain regularization, PAIRED \citep{DJVBRCL:20:NEURIPS} for environment-based opponent modeling, and AdvGAN \citep{XCXLWHML:18:IJCAI} for sample-level adversarial attacks. 
	For \ac{LLM} reasoning, a closely related idea is to make errors explicit in the prompt, as \ac{CoT} demonstrations are typically positive-only. \Ac{CCoT} addresses this by pairing correct and incorrect rationales in the prompt, making errors explicit and improving robustness over vanilla \ac{CoT} \citep{CGTPB:23:ARXIV}. 
	More interactive settings explore stronger counterexamples through debate-style protocols. \citet{AKJHVDRL:24:ICML} finds that more persuasive debaters can lead to more truthful answers, while debate-style adversarial collaboration improves factuality and reduces hallucination \citep{YMFCH:25:AS}. \citet{YZCNLH:25:ICML} studies belief-driven debate/coordination via Bayesian Nash equilibrium to stabilize outcomes. Also, analogical prompting treats \acp{LLM} as analogical reasoners and improves performance by retrieving and reusing solution patterns rather than explicit negatives \citep{MYCXLPJP:24:ICLR}. 
	These lines of work motivate our cycle-based view. Instead of relying on fixed negatives, the bad reasoning should evolve with the solver so that newly exposed weaknesses can be turned into targeted corrections. We also note that our use of adversarial refers to generating plausible-but-wrong reasoning traps and encouraging models to optimize with counterexamples. It should be distinguished from adversarial prompting in \ac{LLM} safety, which typically studies jailbreaks, prompt injection, and harmful instruction elicitation \citep{LDLWWWZLWZ:23:ARXIV,APAZGCAB:25:ICML,SLRAHMFJ:24:NEURIPS,HBCNTF:24:NEURIPS}.

	\subsection{Feedback-Driven Prompt Optimization}
	
	A growing line of work improves \ac{LLM} reasoning by using natural-language feedback to refine outputs or instructions, often without parameter updates. These methods rely on critiques, checks, or structured comments to refine instructions \citep{ZZCLLS:26:ARXIV}. 
	Self-Refine \citep{MSTGHMPC:23:NEURIPS} iteratively generates, critiques, and revises model outputs, achieving consistent gains without additional training. Reflexion \citep{SCGNKY:23:NEURIPS} converts task feedback into short reflective notes that are stored and reused, enabling better future attempts through explicit textual memory rather than weight updates. From the input side, Self-Polish \citep{ZSXZYRGS:23:EMNLP} highlights that refining the problem statement itself can improve multi-step reasoning and robustness, complementing answer-side prompting methods. 
	In addition, prompt-level optimization can be either automated by using \acp{LLM} to search for better instructions (e.g., OPRO \citep{YWLLCZ:24:ICLR}) or guided by human preference feedback when numeric scores are unavailable \citep{XLZDVSNJ:24:ICML}. Also, workflow-level methods can generate and refine multi-step prompting pipelines with feedback signals \citep{JZJYZFTC:25:ICLR}. These approaches support the broader view that iterative, feedback-driven prompt editing can be an effective mechanism for improving reasoning reliability. \ac{CAP-CoT} builds on this view by closing the cycle with an evolving challenger so that feedback improves both the solver's repairs and the challenger's next hard negatives.

	\begin{figure*}[t]
		\centering
			\includegraphics[width=0.9\linewidth]{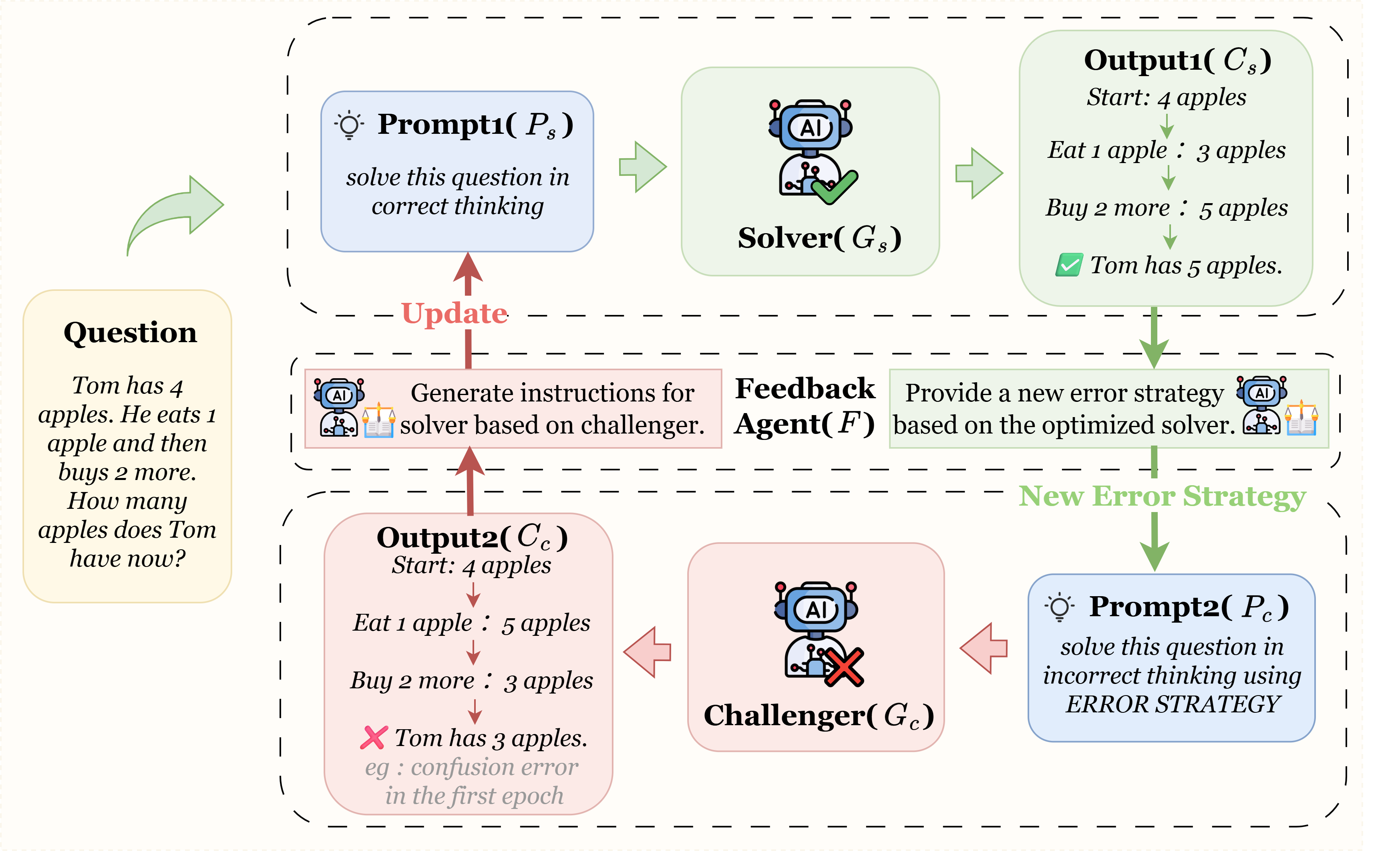}
		\caption{Overview of the proposed framework \ac{CAP-CoT}. }
		\label{fig_main}
	\end{figure*}

	\section{Method}
	\label{sec:3}
	We propose \textbf{\ac{CAP-CoT}}, a {cycle-based} prompt optimization framework for \ac{CoT} reasoning. As shown in Fig.~\ref{fig_main}, \ac{CAP-CoT} uses three role agents, including a solver $G_S$, a challenger $G_C$, and a feedback agent $F$. All roles share the same \ac{LLM} backbone $G_\theta$ and differ in their role prompts and input context. The symbols $G_S$, $G_C$, and $F$ denote prompt-conditioned roles rather than separately trained parameter sets.
	
	The core design is a short optimization \emph{cycle} that repeatedly (i) generates a correct candidate chain, (ii) generates a targeted flawed counter-chain, and (iii) turns their contrast into prompt edits.
	The solver chain is not assumed to be correct during optimization. Instead, it is treated as the current candidate reasoning path produced by the deployed solver. Its final answer is judged by the standard benchmark evaluator only when reporting accuracy, while the feedback agent focuses on local logical validity, missing assumptions, unsupported transitions, and answer-format consistency.
	Each cycle updates the solver prompt to repair fragile reasoning steps and the challenger prompt to produce more diagnostic hard negatives in the next cycle, so the contrast remains informative as the solver improves.
	After a few cycles (typically 2--3), we keep the refined solver prompt and use only the solver at inference time.
	
	\subsection{Solver Agent}
	
	The solver $G_S$, which is the main reasoning model in our framework, aims to produce a reasonable, coherent, and logically consistent \ac{CoT} reasoning sequence given an input query. Let the query be denoted as $Q$, the shared backbone $G_\theta$ produces a forward reasoning chain $C_S$ under a solver prompt $P_S$, which can be formulated as: 
	\begin{align}\label{eq1}
		C_S = G_\theta(P_S + Q).
	\end{align}
	$G_S$ is instructed to generate clearly numbered reasoning steps, state necessary assumptions, separate premises from deductions, and return the final answer in a stable format.
	These constraints reduce avoidable variation across runs by making the solver follow a consistent reasoning schema, and making the chain easier to inspect because later feedback can refer to explicit step indices, even when the underlying reasoning route is imperfect.
	
	\ac{CAP-CoT} does not require the solver to produce a correct chain before refinement. If $C_S$ contains an error, the feedback agent may still identify weak transitions by comparing the chain with the adversarial counter-chain and by checking whether each local inference follows from the stated premises. This avoids using the challenger as an oracle and instead uses the challenger to expose reasoning patterns that the solver should learn to guard against.

	\subsection{Adversarial Challenger Agent}
	
	The challenger $G_C$ is used to construct a \emph{plausible but incorrect} reasoning chain that stands in contrast to the forward reasoning chain $C_S$and stress-tests the current solver prompt $P_S$. Given the same query $Q$ and a challenger prompt $P_C$, the output of $G_C$ is an adversarial chain $C_C$:
	\begin{align}\label{eq2}
		C_C = G_\theta(P_C + Q), 
	\end{align}
	where $G_C$ uses the shared backbone $G_\theta$. The chain $C_C$ follows the surface structure of the task while injecting controlled reasoning flaws, such as logical leaps, conceptual confusions, unsupported assumptions, or locally fluent but globally invalid steps.
	In \ac{CAP-CoT}, \emph{adversarial} refers to \emph{task-semantic} counter-chains that expose logical vulnerabilities in $C_S$, rather than safety-oriented attacks such as jailbreak or prompt injection.
	
	The challenger is designed to be diagnostic rather than merely wrong. A useful $C_C$ should preserve enough correct structure to make the injected flaw comparable with the solver's reasoning. If the adversarial chain is obviously nonsensical, it provides little information for prompt refinement. Also, if it is too similar to the solver chain, it fails to expose a distinct failure mode. The challenger prompt therefore constrains the negative chain to keep the task format, maintain local fluency, and place the error at an identifiable reasoning point.
	
	\noindent\textbf{Cold-Start Error Strategy for the Challenger.}
	We define a lightweight cold-start error taxonomy with four families, \{\emph{jump}, \emph{confusion}, \emph{fuzzy}, \emph{wrapper}\}. The \emph{jump} family omits a necessary intermediate step or validity check. The \emph{confusion} family substitutes a related but incorrect concept, quantity, or condition. The \emph{fuzzy} family hides an inference gap behind vague language, and \emph{wrapper} embeds a wrong core step inside an otherwise fluent explanation. 
	These error types capture common deficiencies observed in reasoning processes, serving as a structured basis for improving robustness and accuracy when facing diverse reasoning tasks. 
	This taxonomy is not intended to cover the full space of \ac{LLM} reasoning errors. It serves as a minimal bootstrap mechanism that initializes the adversarial cycle with common and easily operationalized failure modes. Once feedback becomes available, the challenger prompt is refined toward solver-specific weaknesses and is no longer limited to the original four types.
	Additional analysis of the cold-start error taxonomy and its minimality is provided in Appendix~\ref{app3}.
	
	For each query $Q$, the first cycle samples one or two error families to construct $C_C$, since a single injected flaw is often sufficient for step-level diagnosis, while two families allow interaction effects such as a missing constraint check hidden by fluent wording. We avoid larger combinations because they make the negative chain less interpretable and make it harder for the feedback agent to isolate the causal error.
	In later cycles, the challenger prompt $P_C$ is updated according to feedback so that $G_C$ gradually shifts from generic cold-start errors to targeted hard negatives that match the solver's current blind spots.
	This error strategy makes the contrast between $C_S$ and $C_C$ more informative, and helps $G_C$ maintain challenging attacks as the solver becomes stronger over cycles.

	\subsection{Feedback Agent}
	
	The feedback agent $F$ is responsible for evaluating $C_S$ and $C_C$ and providing structured feedback for both role prompts accordingly. 
	Given $Q$, prompts $(P_S, P_C)$, and reasoning chains $(C_S, C_C)$, $F$ compares the two and produces solver-directed feedback $\mathbb{F}_S$ and challenger-directed feedback $\mathbb{F}_C$:
	\begin{align}\label{eq3}
		(\mathbb{F}_S, \mathbb{F}_C) = G_\theta(P_F, Q, P_S + C_S, P_C + C_C),
	\end{align}
	where $P_F$ is the feedback prompt. 
	$F$ first normalizes both chains into semantic claim units, where each unit contains a local premise, operation, and conclusion. It then aligns units by their role in the solution, such as domain setup, equation transformation, candidate generation, constraint checking, evidence retrieval, or final answer selection. When no one-to-one match exists, the feedback records the unmatched step as a missing, redundant, or alternative reasoning unit rather than forcing a false alignment.
	The feedback follows a compact schema:
	
	\texttt{\{\textbf{Chain}: $C_S$ or $C_C$;}
	
	\texttt{\textbf{Step}: index $t$;}
	
	\texttt{\textbf{Issue type}: missing assumption, incorrect inference, unclear step, etc.;}
	
	\texttt{\textbf{Suggestion}: a short, actionable fix.\}}
	
	To reduce dependence on a noisy challenger, the feedback agent applies two safeguards. Firstly, a solver update is accepted only when the flaw can be stated as a transferable reasoning principle rather than as a task-specific patch. Then, challenger updates are based on the solver's observed weak points rather than on the challenger's final answer alone. These safeguards prevent the optimization loop from overfitting to arbitrary or low-quality negative chains.
	
	\subsection{Structured Feedback Prompt Refinement}
	We adopt an adaptive strategy to refine role \ac{LLM} prompts through a \ac{SFPR} step. 
	In each optimization round, the solver agent $G_S$ generates $C_S$, the adversarial challenger agent $G_C$ generates $C_C$, and the feedback agent $F$ analyzes them jointly.
	The solver prompt is updated as:
	\begin{align}\label{eq4}
		P_S \leftarrow \text{SFPR}(P_S, \mathbb{F}_S).
	\end{align}
	And the challenger prompt is updated as:
	\begin{align}\label{eq5}
		P_C \leftarrow \text{SFPR}(P_C, \mathbb{F}_C),
	\end{align}
	
	\ac{SFPR} consists of four deterministic text-editing stages. The first stage filters feedback entries and keeps only entries that identify a logical or procedural weakness with explicit evidence. The second stage abstracts each retained entry from an instance-specific comment into a reusable directive. For example, a feedback sentence such as ``Step 4 keeps $x=2$ after squaring although it violates the domain'' is rewritten as ``After any transformation that can introduce extraneous candidates, verify each candidate against the original constraints.'' The third stage merges redundant directives by semantic similarity and keeps the most general version. The fourth stage appends at most $K$ concise directives to the corresponding role prompt, where $K$ is fixed before experiments to control prompt growth.
	
	In addition, $F$ keeps its previous records of the latest optimization round termed as $\mathbb{F}_F$. It could refines its own prompt through $P_D \leftarrow \text{SFPR}(P_D, \mathbb{F}_F)$. 
	\ac{SFPR} maintains a transparent refinement cycle and supports an iterative loop with explicit division of labor, where the solver prompt accumulates transferable reasoning safeguards, while the challenger prompt accumulates increasingly targeted probes of unresolved weaknesses.
	
	\subsection{Stopping Criterion and Inference}
	We use a fixed small number of optimization rounds in the main experiments, with three rounds as the default. This choice avoids tuning the stopping point on test performance and keeps the procedure reproducible. We also examine additional rounds in the robustness analysis and observe that most gains appear within the first three rounds. At inference time, \ac{CAP-CoT} uses only the refined solver prompt and a single forward solver call. The challenger and feedback agent are not invoked during final inference, so the deployed inference procedure remains a single-model CoT-style solver.

    \begin{table*}[!t]\small
    	\caption{Performance comparison between our proposed method and representative baseline methods across six reasoning benchmarks under four backbones, with percent symbol (\%) omitted in all the accuracy results. }
    	\label{tab1}
    	\centering
    	\scalebox{1.0}{
    		\begin{tabular}{l|c|ccccccc}
    			\toprule
    			\textbf{Methods} & \textbf{Backbone} & \textbf{MATH} & \textbf{GSM8K} & \textbf{BBH} & \textbf{MMLU-CF} & \textbf{HotpotQA} & \textbf{LongBench} & \textbf{Avg.} \\ 
    			\midrule
    			\midrule
    			\ac{CoT}~\citep{KGSRMYI:22:NEURIPS} & \multirow{14}{*}{GPT-4o-mini} & 78.3  & 90.9  & 78.3  & 69.6  & 67.2  & 57.6  & 73.6  \\
    			\acs{CoT-SC}~\citep{XWSQVLEN:23:ICLR} & & 81.8  & 92.0  & 83.4  & 71.1  & 66.2  & 58.6  & 75.5  \\
    			Self-Refine~\citep{MSTGHMPC:23:NEURIPS} & & 78.7  & 91.7  & 80.0  & 69.7  & 68.3  & 58.2  & 74.4  \\
    			\ac{PAgent}~\citep{XWCZWFBH:24:ICLR} & & 80.6 & 92.5 & 80.1 & 70.5 & 69.8 & 58.7 & 75.4 \\
    			\acs{AP}~\citep{MYCXLPJP:24:ICLR}  & & 65.4  & 87.2  & 72.5  & 65.8  & 64.7  & 52.9  & 68.1  \\
    			\ac{CCoT}~\citep{CGTPB:23:ARXIV} & & 80.1 & 92.4 & 80.5 & 70.4 & 71.5 & 58.8 & 75.6 \\
    			\acs{AFlow}~\citep{JZJYZFTC:25:ICLR}  & & 83.0  & 93.5  & 76.0  & 69.5  & 73.5  & 61.0  & 76.1  \\
    			\ac{ToT}~\citep{YDZGTGCN:23:NEURIPS} & & 83.0  & 94.5  & 85.9  & 70.2  & 74.2  & 63.1  & 78.5  \\
    			\ac{GoT}~\citep{BMBKRGMP:24:AAAI} & & 82.3  & 94.9  & 84.1  & 71.6  & 76.8  & 63.8  & 78.9  \\
    			MAD~\citep{AKJHVDRL:24:ICML} & & 82.9  & 94.6  & 85.2  & 71.0  & 75.5  & 64.2  & 78.9  \\
    			ECON~\citep{YZCNLH:25:ICML}   & & 83.5  & 95.0  & 86.0  & 71.5  & 77.0  & 65.0  & 79.7  \\
    			\ac{FoT}~\citep{BHLTW:25:ICML} & & 82.5  & 94.0  & 82.4  & 70.6  & 66.7  & 59.1  & 76.1  \\
    			\ac{AoT}~\citep{TYSZLW:26:NeurIPS} & & 83.6  & 95.0  & 86.0  & 70.9  & 80.6  & 68.5  & 80.8  \\
    			\ac{CAP-CoT} (Ours)   & & \textbf{87.2} & \textbf{96.1} & \textbf{87.9} & \textbf{72.5} & \textbf{83.1} & \textbf{69.3} & \textbf{82.7} \\ \midrule
    			
    			\ac{CoT}~\citep{KGSRMYI:22:NEURIPS} & \multirow{14}{*}{Qwen-turbo} & 78.1  & 90.7  & 78.1  & 69.4  & 66.8  & 57.3  & 73.4  \\
    			\acs{CoT-SC}~\citep{XWSQVLEN:23:ICLR} & & 81.4  & 91.5  & 83.2  & 70.8  & 65.9  & 58.4  & 75.2  \\
    			Self-Refine~\citep{MSTGHMPC:23:NEURIPS} & & 78.5  & 91.4  & 79.8  & 69.5  & 68.2  & 58.0  & 74.2  \\
    			\ac{PAgent}~\citep{XWCZWFBH:24:ICLR} & & 80.1 & 92.2 & 80.4 & 70.1 & 68.9 & 58.5 & 75.0 \\
    			\acs{AP}~\citep{MYCXLPJP:24:ICLR}  & & 65.2  & 87.0  & 72.2  & 65.2  & 64.5  & 52.7  & 67.8  \\
    			\ac{CCoT}~\citep{CGTPB:23:ARXIV} & & 79.9 & 92.1 & 80.2 & 70.1 & 71.2 & 58.5 & 75.3 \\
    			\acs{AFlow}~\citep{JZJYZFTC:25:ICLR}  & & 82.4  & 93.1  & 75.7  & 69.3  & 73.2  & 60.4  & 75.7  \\
    			\ac{ToT}~\citep{YDZGTGCN:23:NEURIPS} & & 81.9  & 94.2  & 83.7  & 71.3  & 76.4  & 62.4  & 78.3  \\
    			\ac{GoT}~\citep{BMBKRGMP:24:AAAI} & & 82.7  & 93.8  & 84.9  & 70.1  & 74.0  & 62.8  & 78.1  \\
    			MAD~\citep{AKJHVDRL:24:ICML} & & 83.2  & 93.8  & 83.5  & 70.8  & 74.5  & 63.2  & 78.2  \\
    			ECON~\citep{YZCNLH:25:ICML}   & & 83.9  & 94.2  & 84.4  & 71.2  & 75.8  & 64.0  & 78.9  \\
    			\ac{FoT}~\citep{BHLTW:25:ICML} & & 82.2  & 93.9  & 82.3  & 70.4  & 66.4  & 59.0  & 75.7  \\
    			\ac{AoT}~\citep{TYSZLW:26:NeurIPS} & & 83.5  & 94.7  & 85.4  & 70.5  & 77.5  & 68.1  & 79.9  \\
    			\ac{CAP-CoT} (Ours)   & & \textbf{87.2} & \textbf{95.6} & \textbf{87.1} & \textbf{71.6} & \textbf{80.4} & \textbf{68.9} & \textbf{81.8} \\ \midrule
    			
    			\ac{CoT}~\citep{KGSRMYI:22:NEURIPS} & \multirow{14}{*}{Deepseek-V3} & 78.5  & 91.3  & 78.5  & 69.9  & 67.4  & 57.7  & 73.9  \\
    			\acs{CoT-SC}~\citep{XWSQVLEN:23:ICLR} & & 82.0  & 92.1  & 83.6  & 71.5  & 66.6  & 58.9  & 75.8  \\
    			Self-Refine~\citep{MSTGHMPC:23:NEURIPS} & & 78.9  & 91.9  & 80.4  & 70.1  & 69.1  & 58.4  & 74.8  \\
    			\ac{PAgent}~\citep{XWCZWFBH:24:ICLR} & & 80.8 & 92.8 & 81.2 & 70.6 & 69.7 & 59.2 & 75.7 \\
    			\acs{AP}~\citep{MYCXLPJP:24:ICLR}  & & 65.6  & 87.6  & 72.8  & 66.1  & 64.9  & 53.4  & 68.4  \\
    			\ac{CCoT}~\citep{CGTPB:23:ARXIV} & & 80.4 & 92.1 & 80.2 & 70.1 & 71.2 & 58.5 & 75.3 \\
    			\acs{AFlow}~\citep{JZJYZFTC:25:ICLR}  & & 83.4  & 93.6  & 76.4  & 69.8  & 74.0  & 61.4  & 76.4  \\
    			\ac{ToT}~\citep{YDZGTGCN:23:NEURIPS} & & 82.5  & 95.0  & 84.4  & 72.0  & 76.9  & 63.2  & 79.0  \\
    			\ac{GoT}~\citep{BMBKRGMP:24:AAAI} & & 83.2  & 94.5  & 86.2  & 70.3  & 74.2  & 63.4  & 78.6  \\
    			MAD~\citep{AKJHVDRL:24:ICML} & & 84.1  & 94.5  & 84.8  & 71.2  & 76.8  & 64.5  & 79.3  \\
    			ECON~\citep{YZCNLH:25:ICML}   & & 84.8  & 94.9  & 85.5  & 71.8  & 78.0  & 65.2  & 80.0  \\
    			\ac{FoT}~\citep{BHLTW:25:ICML} & & 82.7  & 94.2  & 82.6  & 70.5  & 66.8  & 59.3  & 76.0  \\
    			\ac{AoT}~\citep{TYSZLW:26:NeurIPS} & & 84.0  & 95.1  & 86.1  & 70.8  & 80.6  & 68.7  & 80.9  \\
    			\ac{CAP-CoT} (Ours)   & & \textbf{87.5} & \textbf{96.0} & \textbf{87.8} & \textbf{72.2} & \textbf{82.8} & \textbf{69.5} & \textbf{82.6} \\ \midrule
    			
    			\ac{CoT}~\citep{KGSRMYI:22:NEURIPS} & \multirow{14}{*}{GPT-4o} & 79.5  & 92.5  & 79.8  & 71.5  & 69.5  & 59.5  & 75.4  \\
    			\acs{CoT-SC}~\citep{XWSQVLEN:23:ICLR} & & 82.8  & 93.2  & 84.5  & 72.8  & 68.2  & 60.8  & 77.1  \\
    			Self-Refine~\citep{MSTGHMPC:23:NEURIPS} & & 80.2  & 92.8  & 81.6  & 71.8  & 71.5  & 60.2  & 76.4  \\
    			\ac{PAgent}~\citep{XWCZWFBH:24:ICLR} & & 82.1 & 93.9 & 82.5 & 72.3 & 72.2 & 61.0 & 77.3 \\
    			\acs{AP}~\citep{MYCXLPJP:24:ICLR}  & & 67.5  & 88.5  & 74.2  & 67.5  & 66.8  & 55.2  & 70.0  \\
    			\ac{CCoT}~\citep{CGTPB:23:ARXIV} & & 81.5 & 93.8 & 81.9 & 72.1 & 73.8& 60.5 & 77.3  \\
    			\acs{AFlow}~\citep{JZJYZFTC:25:ICLR}  & & 84.5  & 94.8  & 77.5  & 71.2  & 75.5  & 63.5  & 77.8  \\
    			\ac{ToT}~\citep{YDZGTGCN:23:NEURIPS} & & 83.5  & 95.5  & 85.5  & 73.5  & 78.5  & 64.5  & 80.2  \\
    			\ac{GoT}~\citep{BMBKRGMP:24:AAAI} & & 84.2  & 95.2  & 87.0  & 72.2  & 76.2  & 65.2  & 80.0  \\
    			MAD~\citep{AKJHVDRL:24:ICML} & & 85.8  & 95.5  & 86.2  & 73.0  & 78.5  & 65.2  & 80.7  \\
    			ECON~\citep{YZCNLH:25:ICML}   & & 86.4  & 95.9  & 86.9  & 73.5  & 79.8  & 66.0  & 81.4  \\
    			\ac{FoT}~\citep{BHLTW:25:ICML} & & 83.8  & 94.9  & 83.5  & 72.4  & 68.5  & 61.5  & 77.4  \\
    			\ac{AoT}~\citep{TYSZLW:26:NeurIPS} & & 85.5  & 95.8  & 87.2  & 73.0  & 81.5  & 69.5  & 82.1  \\
    			\ac{CAP-CoT} (Ours)   & & \textbf{89.2} & \textbf{96.6} & \textbf{89.1} & \textbf{74.5} & \textbf{84.0} & \textbf{70.4} & \textbf{84.0} \\
    			\bottomrule
    		\end{tabular}
    	}
    \end{table*}

	\section{Experiments}
	\label{sec:4}	
	\subsection{Experimental Setup}
	\noindent \textbf{Datasets and Evaluation}: 
	We evaluate our method on six \ac{QA} benchmarks: MATH \citep{HBKABTSS:21:NEURIPS}, GSM8K \citep{CKB:21:ARXIV}, BBH \citep{SSSGCCTZ:23:ACL}, MMLU-CF \citep{ZHYLCMXYYL:24:ARXIV}, HotpotQA \citep{YZSZBYCR:18:EMNLP}, and LongBench \citep{BZLJTHZD:24:ACL}. Collectively, these datasets cover advanced and grade-school math, logical and algorithmic reasoning, commonsense and factual knowledge, multi-hop \ac{QA}, and very long-context understanding.
	For all experiments, we adopt standard accuracy as the evaluation metric. The optimization agents do not receive the gold answer during prompt refinement; gold answers are used only by the evaluator to compute final accuracy.
	
	\medskip
	\noindent \textbf{Baselines}: 
	We compare \ac{CAP-CoT} with a broad set of strong reasoning baselines, including standard \ac{CoT} \citep{KGSRMYI:22:NEURIPS}; \ac{CoT-SC} \citep{XWSQVLEN:23:ICLR}, which samples multiple reasoning paths ($n\!=\!5$) and selects the most consistent answer; \ac{CCoT} \citep{CGTPB:23:ARXIV} that enhance reasoning demonstration; enhanced CoT-like methods including \ac{ToT} \citep{YDZGTGCN:23:NEURIPS}, \ac{GoT} \citep{BMBKRGMP:24:AAAI}, \ac{FoT} \citep{BHLTW:25:ICML} (with reasoning branches $n\!=\!8$), and \ac{AoT} \citep{TYSZLW:26:NeurIPS}; {Self-Refine} \citep{MSTGHMPC:23:NEURIPS}, where an \ac{LLM} iteratively generates, critiques, and revises its own outputs; prompt-optimization or workflow-optimization baselines such as \ac{PAgent} \citep{XWCZWFBH:24:ICLR} and \ac{AP} \citep{MYCXLPJP:24:ICLR}; feedback-driven approaches such as \ac{AFlow} \citep{JZJYZFTC:25:ICLR}; and debate- or committee-style methods such as MAD \citep{AKJHVDRL:24:ICML} and ECON \citep{YZCNLH:25:ICML}.
	
	\medskip
	\noindent \textbf{Implementation Details}:
	\ac{CAP-CoT} uses three role agents on a shared \ac{LLM} backbone: the solver $G_S$, the adversarial challenger $G_C$, and the feedback agent $F$. For backbone comparison, we instantiate all three roles with GPT-4o-mini, Qwen-turbo, DeepSeek-V3, and GPT-4o. Unless otherwise stated, \emph{GPT-4o-mini} is used as the default backbone for its good trade-off between reasoning ability and latency. We implement the framework in LangChain, which lets us modularize the three roles and call \acp{LLM} via official APIs. Unless otherwise noted, we disable nucleus sampling, set the maximum generation length to 2048 tokens, and set both frequency and presence penalties to 0.0. 
	Except for robustness-to-rounds experiments, all reported results are averaged over three independent runs; ablation studies use five runs with the same averaging protocol. 
	In accuracy-focused ablations, the temperature is fixed at 0.0, while in robustness-to-rounds experiments, we sweep the temperature from 0.0 to 1.0 on the MATH dataset.
	For reproducibility, example role prompts for the solver, challenger, and feedback agent are provided in Appendix~\ref{app1}.

	\subsection{Performance Evaluation}
	
	Table \ref{tab1} compares \ac{CAP-CoT} with all baselines on six benchmarks under four \ac{LLM} backbones, using results of \emph{three} rounds of the optimization cycle for our method. \ac{CAP-CoT} consistently outperforms all baselines across all datasets under four \ac{LLM} backbones, and achieves its strongest results on the most capable backbone, GPT-4o. It shows clear gains on math-heavy and multi-step reasoning tasks, where errors often arise from missing constraints, multi-step dependency failures, evidence-selection mistakes, or long-context distraction. These are precisely the cases in which a static forward chain can look locally plausible while remaining globally fragile.
	
	While \ac{CoT} and its variants perform reasonably well on simpler problems, they lag behind on multi-hop and long-context \ac{QA}, and \ac{AP} is generally weaker, suggesting that analogy alone is not sufficient for complex reasoning. In contrast, our proposed framework yields more accurate, stable, and backbone-agnostic solvers.
	Compared with \ac{PAgent}, which also uses error feedback for prompt optimization, \ac{CAP-CoT} differs in how the optimization signal is formed. \ac{PAgent} searches over prompt candidates guided by task-level feedback, whereas \ac{CAP-CoT} constructs a task-semantic hard negative and converts the contrast between the solver chain and the adversarial chain into step-level prompt edits. The empirical advantage of \ac{CAP-CoT} suggests that the evolving negative chain provides additional diagnostic information beyond scalar correctness or generic error feedback.
	
	The results also show that \ac{CAP-CoT} transfers across backbones with different baseline strengths, suggesting that the method mainly improves prompt-level reasoning behavior rather than exploiting idiosyncrasies of a single API model. 
	However, because the evaluated backbones are not specialized long-form reasoning models with explicit hidden thinking traces, we do not claim that these results fully characterize transfer to reasoning-oriented models.

	\subsection{Ablation Studies}
	
	\subsubsection{Effect of Framework Components}
	
	\begin{table}[!t]
		\caption{Ablation results of different components in our proposed framework on the MATH dataset.}
		\label{tab_abl}
		\centering
		\scalebox{0.85}{
			\begin{tabular}{lcc}
				\toprule
				Model Components & Accuracy(\%) & Mean Variation \\
				\midrule
				\midrule
				$G_S$ & 82.5 & 4.15 \\
				$G_S+F$ & 83.8 & 3.80 \\
				$G_S+G_C+F$ & 87.2 & 2.62 \\
				\bottomrule
			\end{tabular}
		}
	\end{table}
	
	Table \ref{tab_abl} shows the ablation results assessing the contribution of each module on MATH dataset. Using only the solver $G_S$ gives a strong baseline, while adding the feedback agent $F$ already improves accuracy and slightly reduces mean variation, indicating that structured critique alone helps the solver tighten its reasoning. Introducing the adversarial challenger $G_C$ on top of this ($G_S{+}G_C{+}F$) brings the largest gain, with the highest accuracy and the lowest variation.
	This result supports the central design of \ac{CAP-CoT}, where feedback becomes more useful when it is grounded in a plausible contrastive failure rather than produced from the solver chain alone.
	Introducing the adversarial challenger $G_C$ on top of this ($G_S+G_C+F$) brings the largest gain, with the highest accuracy and the lowest variation.
	This result supports the central design of CAP-CoT, where feedback becomes more useful when it is grounded in a plausible contrastive failure rather than produced from the solver chain alone.
	This suggests that the challenger contributes by providing informative contrastive failures rather than merely adding noise to the refinement process.
	Qualitative examples illustrating how the solver, challenger, and feedback agent interact during one optimization cycle are provided in Appendix~\ref{app2}.

	\subsubsection{Performance over Optimization Rounds}
	
	\begin{figure*}[!t]
	    \centering
	    \includegraphics[width=0.9\linewidth]{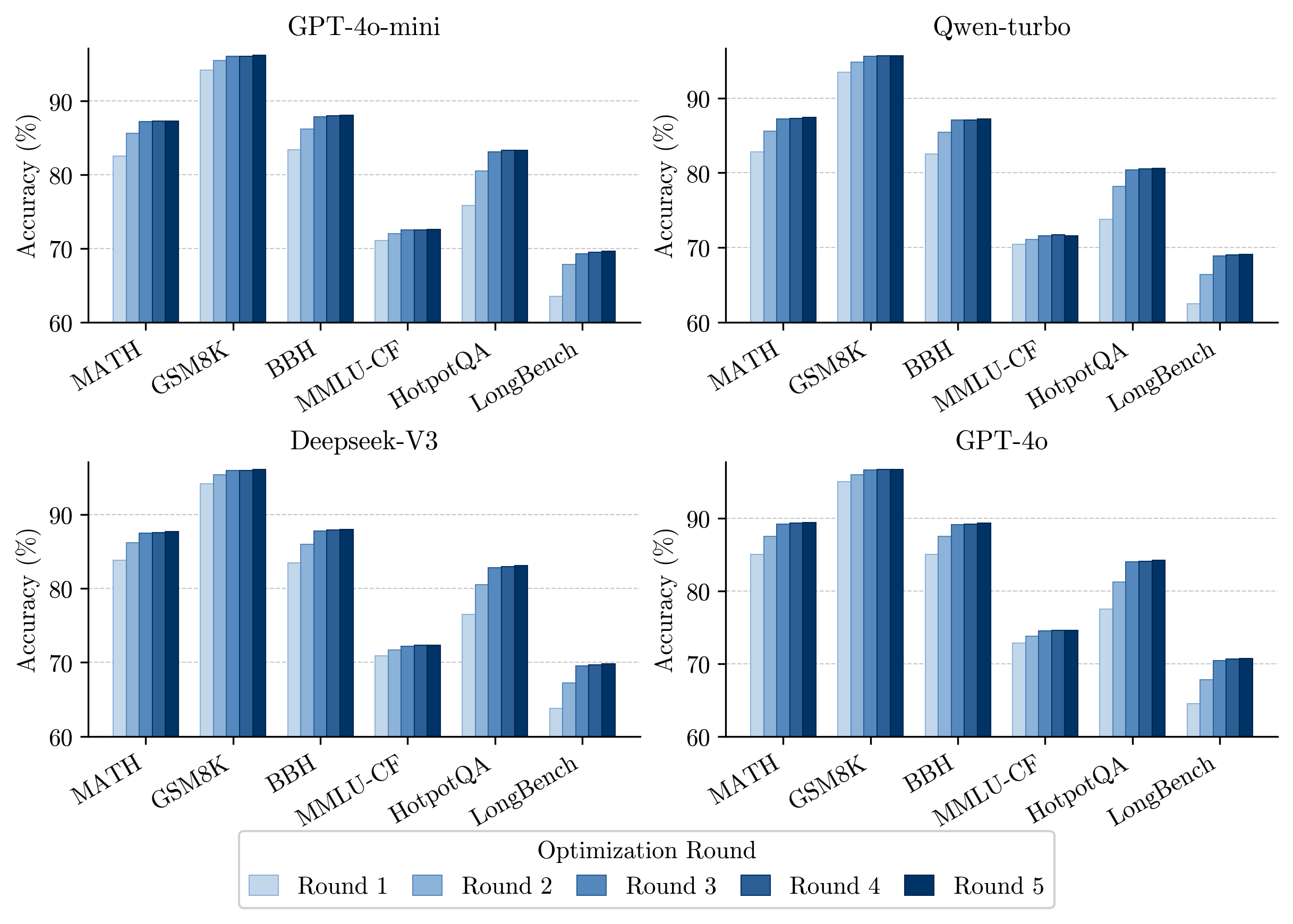}
	    \caption{Accuracy results (\%) of our framework over optimization rounds on four \ac{LLM} backbones.}
	    \label{fig_round}
	\end{figure*}
	
	We further present the performance of \ac{CAP-CoT} across different optimization rounds in Figure \ref{fig_round}. Round 1 denotes the solver after one full cycle of adversarial chain generation and feedback-based refinement, not the raw zero-cycle solver. Across all backbones and benchmarks, reasoning accuracy generally improves from round 1 to round 3, and later gains become smaller. This saturation pattern explains why we use \emph{three rounds} for the main results. A small fixed round budget also limits prompt growth, reduces optimization cost, and avoids selecting the stopping point based on test-set performance. The round-wise trend further clarifies the role of the adaptive challenger. Early cycles expose broad reasoning weaknesses, such as skipped constraints or concept confusion, and later cycles focus on narrower errors tied to the current solver prompt.

	\subsubsection{Effect of Hyperparameter}
	
	\begin{figure*}
		\centering
		\includegraphics[width=0.9\linewidth]{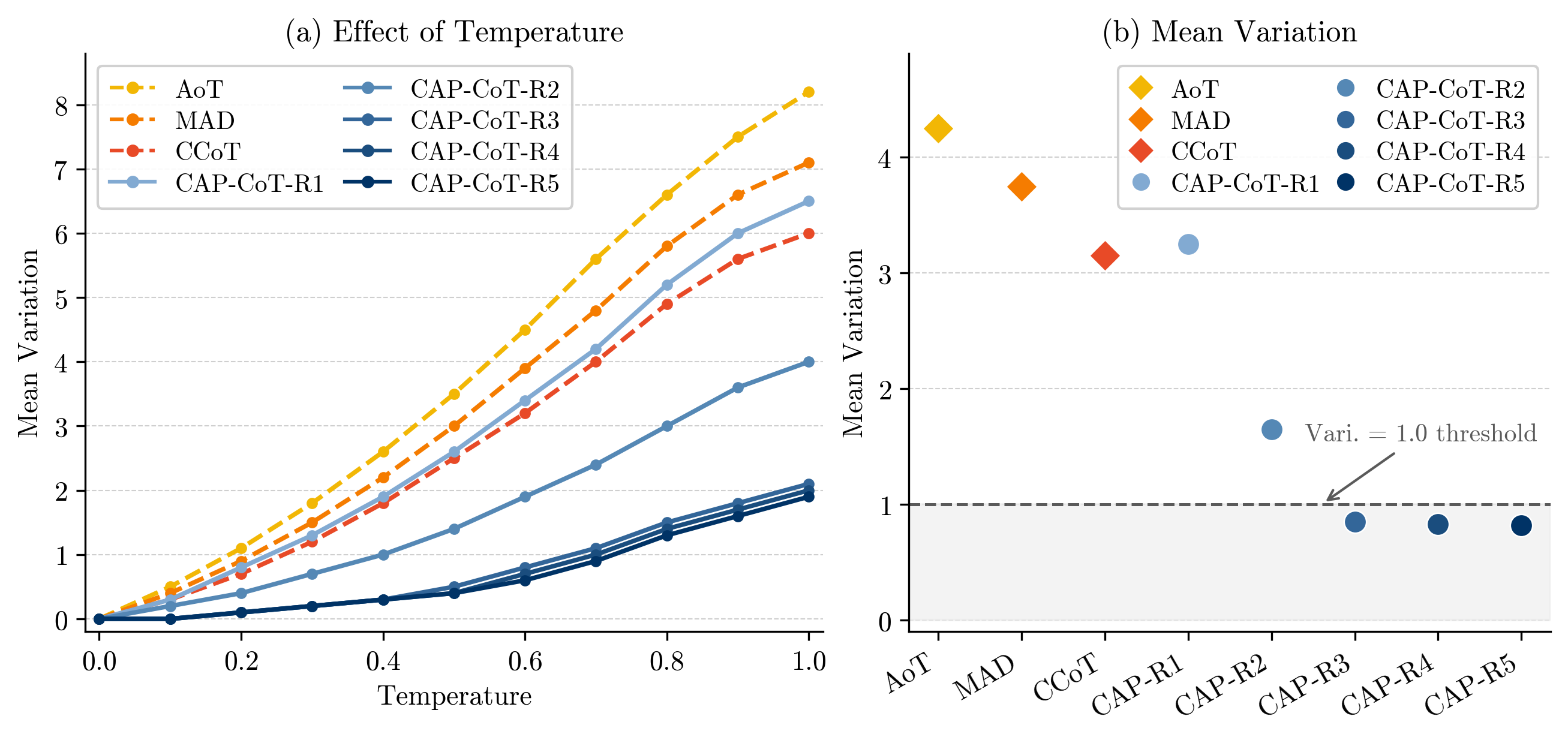}
		\caption{Effect of LLM temperature on reasoning stability on the MATH dataset. Left: Accuracy variation as a function of temperature for three baselines and CAP-CoT under different optimization rounds. Right: Mean variation of each method aggregated over the tested temperature range, with the dashed line marking the variation threshold of 1.0.}
		\label{fig_llmtemp}
	\end{figure*}

	Fig.~\ref{fig_llmtemp} examines how temperature affects reasoning stability on MATH for \ac{AoT}, MAD, \ac{CCoT}, and our method across optimization rounds, with temperatures ranging from 0 to 1 in increments of 0.1. 
	As shown in the left subplot, for all baselines, variation grows quickly as temperature increases, with relatively high mean variation. 
	Our method starts at a similar level in round~1, but successive optimization rounds substantially reduce sensitivity to temperature: the mean variation drops below 1.0 by round~3, as summarized in the right subplot, and continues to decrease slightly thereafter. 
	At higher temperatures (e.g., 0.7--1.0), our optimized solver remains much more stable than \ac{AoT}, MAD, and \ac{CCoT}, indicating that adversarial optimization with structured feedback not only improves accuracy but also makes the model more robust to sampling noise.

	\subsection{Computational Efficiency Analysis}
	
	\begin{figure*}
		\centering
		\includegraphics[width=0.9\linewidth]{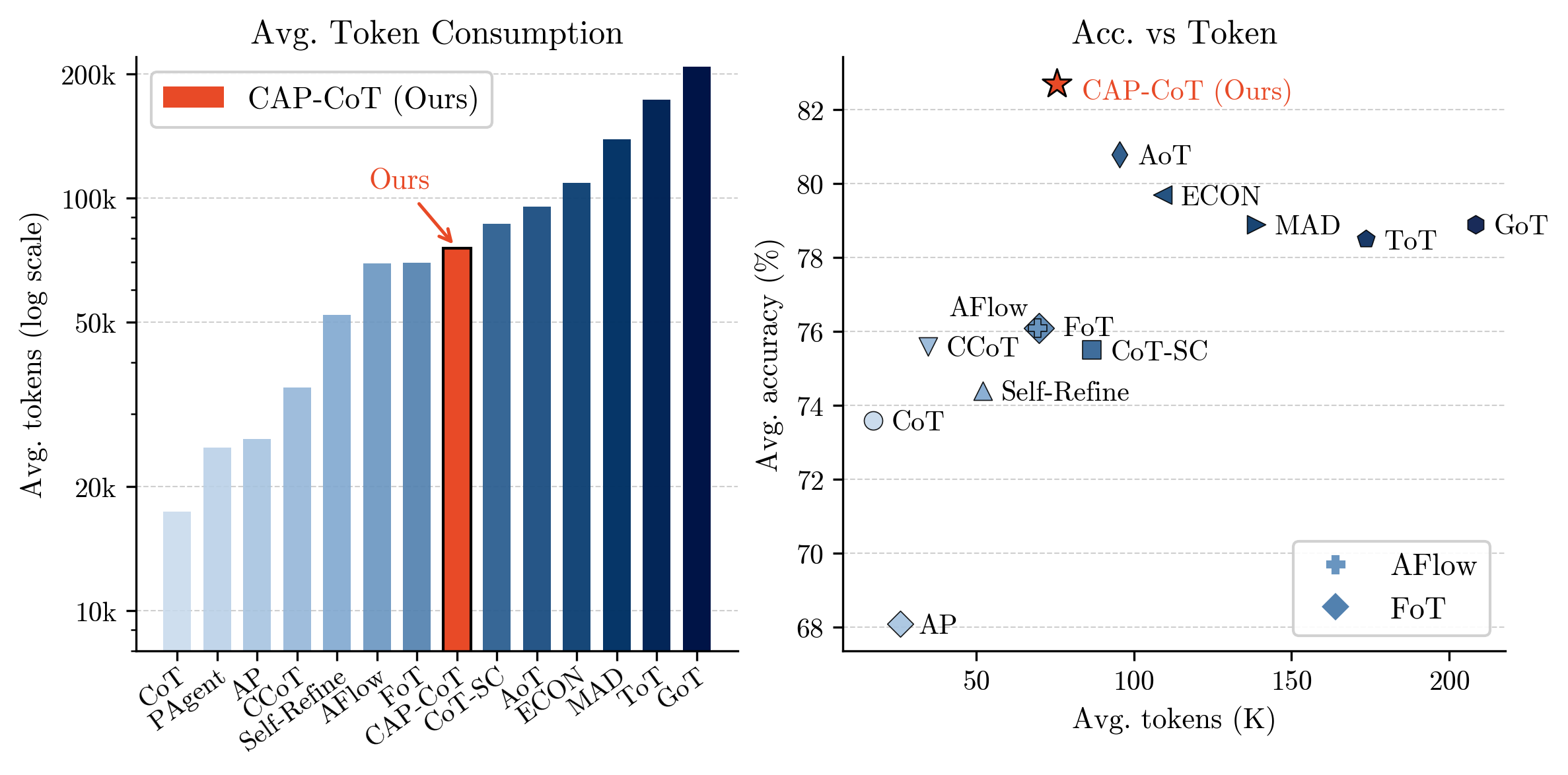}
		\caption{Comparison of token consumption and accuracy--cost trade-off across six reasoning benchmarks. Left: Average token consumption per question for different reasoning methods, with CAP-CoT highlighted in red. Right: Average accuracy as a function of average token consumption, illustrating the trade-off between reasoning performance and token cost.}
		\label{fig_token}
	\end{figure*}
	
	Fig.~\ref{fig_token} presents the token usage of \ac{CAP-CoT} compared with all baselines across six datasets. 
	As shown in the left subplot, single-pass methods such as \ac{CoT} and \ac{AP} are the most cost-efficient, since they generate only one reasoning trace per query. In contrast, methods that rely on explicit test-time scaling, either by sampling and voting such as \ac{CoT-SC} and \ac{FoT}, structured search over many branches (\ac{ToT}, \ac{GoT}), or multi-turn debating and coordination (MAD, ECON, \ac{AoT}), require substantially higher token costs, often exceeding \ac{CoT} by a large margin due to repeated generation, evaluation, and aggregation. \ac{CAP-CoT} sits in the middle of this spectrum, as its cost is higher than simple CoT-style baselines because each optimization cycle involves additional generations from the challenger and feedback roles. But, \ac{CAP-CoT} remains markedly cheaper than heavy tree-, graph-, and debate-based methods, since it does not expand into a large branching space at inference time, nor does it require long multi-round deliberation for every test query. 
	This indicates that our cycle-based adversarial optimization provides a favorable trade-off between performance and token consumption.

	\section{Conclusion}
	\label{sec:5}
	In this work, we propose \ac{CAP-CoT}, a cycle adversarial prompt optimization framework for improving the accuracy and stability of reasoning chain reasoning. \ac{CAP-CoT} pairs a solver with a challenger agent that generates targeted erroneous reasoning chains to contrast the solver's reasoning chain, and a feedback agent that compares the two chains to produce step-level feedback. This feedback refines the solver prompt to strengthen fragile reasoning steps and adapts the challenger prompt to generate more targeted future attacks, forming an iterative optimization cycle while keeping inference as a single-model setup. 
	Extensive experiments on six benchmarks under four \ac{LLM} backbones show that \ac{CAP-CoT} consistently outperforms strong reasoning baselines. Ablations further confirm the contributions of the challenger and feedback agent, and demonstrate improved robustness and stability with a favorable token-cost trade-off.

	\section{Acknowledgements}
	This work was supported by the National Research Foundation of Korea (NRF) grant funded by the Korean government (MSIT) under RS-2025-00556064; by the MSIT (Ministry of Science and ICT), Korea, under the ITRC (Information Technology Research Center) support program (IITP-2026-RS-2021-II212046) supervised by the IITP (Institute for Information \& Communications Technology Planning \& Evaluation); and by the National Natural Science Foundation of China (NSFC) under the General Program (Grant No.\ 62572104).

	\bibliographystyle{cas-model2-names}
	\bibliography{ref}
	
	\appendix
	\renewcommand{\thetable}{\thesection.\arabic{table}}
	\setcounter{table}{0}
	
	\section{Appendix A: Examples Prompts for Role Agents}
	\label{app1}
	This appendix provides example prompts used in \ac{CAP-CoT} for the three roles in one optimization cycle: the Solver ($G_S$), the Challenger ($G_C$), and the Feedback Agent ($F$). All roles may share the same \ac{LLM} backbone but use different role instructions. For clarity and reproducibility, we keep unified placeholders (e.g., $\{Input\ Question\}$, $\{Strategy\ Definition\}$) that are filled with instance-specific content at runtime.
	
	\subsection{Solver Prompt ($G_S$)}
	
	\noindent\textbf{Role Definition.} You are an expert reasoning engine designed to solve complex problems with high accuracy and stability. Your goal is to derive the correct answer through a rigorous, step-by-step \ac{CoT} process.
	
	\noindent\textbf{Base Instructions.}
	(1) Analyze the request: identify the core question, key variables, and constraints.
	(2) Step-by-step derivation: break the problem into logical sub-steps; for each step, explicitly state the premise and conclusion.
	(3) Self-verification: briefly check the logic of each step before moving to the next to prevent error propagation.
	(4) Final answer: conclude with a clear and concise final answer.
	
	\noindent\textbf{Dynamic Guidelines (Iteratively Updated).} This section is empty in the initial cycle. In later cycles, it is populated with specific, context-aware constraints produced by the feedback agent to address weaknesses revealed by the contrast. A placeholder example is: ``Define all variables before use, and verify unit consistency in the final computation.''
	
	\noindent\textbf{Input Task.}
	Question: $\{Input\ Question\}$.
	Instruction: Provide your reasoning chain and final answer.
	
	\subsection{Challenger Prompt ($G_C$)}
	
	\noindent\textbf{Role Definition.} You are an adaptive adversarial challenger. Your goal is \emph{not} to solve the problem correctly. Instead, generate a plausible but incorrect reasoning chain that serves as a hard negative sample to test the solver's robustness.
	
	\noindent\textbf{Core Objective.} Construct a reasoning chain that matches the style and tone of a correct solution but contains a specific flaw dictated by the adversarial instruction. The error should be subtle enough to mislead a careless solver, yet logically fatal to the final answer.
	
	\noindent\textbf{Adversarial Instruction (Input).} This instruction specifies the error type you must inject. It can be a predefined category or a context-aware directive.
	
	\noindent\textbf{Strategy Definition (Context).} Detailed constraints for the current strategy are provided below. You must follow them when constructing the error.
	Definition: $\{Strategy\ Definition\}$.
	
	\noindent\textbf{Execution Guidelines.}
	(1) Plausibility is key: avoid obvious nonsense; keep a high-quality step-by-step structure.
	(2) Targeted sabotage: inject the error only as required by $\{Strategy\ Definition\}$; keep the rest coherent to make the flaw hard to spot.
	(3) Incorrect conclusion: ensure the reasoning leads to a final answer that is wrong and distinct from the ground truth.
	
	\noindent\textbf{Input Task.}
	Question: $\{Input\ Question\}$.
	Instruction: Generate the adversarial reasoning chain following the strategy definition above.
	
	\subsection{Feedback Agent Prompt ($F$)}
	
	\noindent\textbf{Role Definition.} You are the meta-optimization controller. Your objective is to drive the solver ($G_S$) toward a logically flawless Chain-of-Thought. You do this by running an evolutionary loop with the challenger ($G_C$). You must execute a strict two-step process: first, extract high-level reasoning principles to strengthen the solver; second, design a new adversarial strategy to stress-test the solver from a different angle in the next cycle.
	
	\noindent\textbf{Input Context.}
	(1) Question: the problem statement.
	(2) Challenger output ($C_C$): the adversarial reasoning chain (negative sample).
	(3) Solver output ($C_S$): the solver's reasoning chain (target sample).
	
	\noindent\textbf{Task Instructions (Two-Step Process).}
	
	\noindent\textbf{Step 1: Comparative Analysis and Solver Improvement (Optimizing $G_S$).}
	Dissect $C_C$ to identify the fundamental logical flaw or structural gap it exploits. Assess whether $C_S$ is robust enough to prevent this kind of flaw; even if $C_S$ is correct, identify weaknesses in rigor, clarity, or verification. Then synthesize a high-level improvement principle for $G_S$. The goal is to raise the solver's overall reasoning standard rather than patching a single case.
	
	\noindent\textbf{Step 2: Strategy Diversification (Directing $G_C$).}
	Assume the solver will adapt to the previous error type. Choose a distinct, unexplored dimension of reasoning to test next, and formulate a new adversarial strategy that targets a potential weakness in $C_S$. This directive will be used to populate the challenger's ``Strategy Definition'' in the next cycle.
	
	\noindent\textbf{Output Format.}
	
	\noindent[Step 1: Comparative Logic Analysis]
	Adversarial Logic Flaw: [technical description of the flaw in $C_C$].
	Solver Logic Assessment: [evaluation of $C_S$'s robustness regarding this flaw].
	
	\noindent[Step 1 Output: Logical Enhancement Directive for $G_S$]
	Imperative principle to elevate reasoning rigor:
	[directive for $G_S$, e.g., ``Enforce a strict premise--deduction--conclusion structure for each step, and add minimal checks for key assumptions.'']
	
	\noindent[Step 2: Strategic Rationale]
	Evolutionary Direction: [why the new adversarial focus is distinct and necessary].
	
	\noindent[Step 2 Output: Next-Step Adversarial Strategy for $G_C$]
	Strategy Name: [formal strategy designation].
	Strategy Definition: [precise directive for the next negative sample, e.g., ``Keep arithmetic correct but misinterpret the direction of a key constraint, and ensure the mistake appears at an early step.'']
	
	\noindent\textbf{Structured Feedback Prompt Refinement.} \ac{CAP-CoT} uses a unified \ac{SFPR} mechanism to convert the Feedback Agent's outputs into prompt updates for both the Solver and the Challenger. \ac{SFPR} works as follows: it first extracts the high-level reasoning principle from the feedback, removes instance-specific details while keeping transferable constraints, rewrites them into concise imperative natural language, de-duplicates repeated items and enforces a maximum number of updates, and finally appends the resulting instructions to the corresponding role prompt. 

	\begin{table*}[ht]
		\centering
		\small
		\setlength{\tabcolsep}{6pt}
		\renewcommand{\arraystretch}{1.1}
		\begin{tabular}{p{0.97\linewidth}}
			\toprule
			\textbf{Example B.1: Jump-type hard negative for enforcing constraint checks (MATH-style).}\\ 
			\midrule
			\midrule
			\textbf{Question.} Solve $\sqrt{x-1} = x-3$ over the reals. \\
			\textbf{Solver output $C_S^{(0)}$.}
			(1) Domain: require $x-1 \ge 0$ and $x-3 \ge 0$, so $x \ge 3$.
			(2) Square both sides: $x-1 = (x-3)^2 = x^2 - 6x + 9$.
			(3) Rearrange: $0 = x^2 - 7x + 10 = (x-5)(x-2)$.
			(4) Candidates: $x=2$ or $x=5$.
			(5) \emph{Answer: ${2,5}$.} \\
			\textbf{Challenger output $C_C^{(0)}$ (Jump error: skip the domain filter).}
			(1) Square both sides to remove the root: $x-1 = (x-3)^2$.
			(2) Solve $(x-5)(x-2)=0$, so $x=2$ or $x=5$.
			(3) \emph{Answer: ${2,5}$.} \\
			\textbf{Feedback (step-aligned).}
			\textbf{For $C_S^{(0)}$:} Step (4)--(5) retains candidates without applying the domain constraint $x \ge 3$. Require an explicit \emph{candidate $\rightarrow$ constraint check $\rightarrow$ substitution check} pattern before concluding.
			\textbf{For $C_C^{(0)}$:} The chain hides the key validity condition by omitting domain filtering, making the extra root appear legitimate. \\
			\textbf{Solver output $C_S^{(1)}$ (after one \ac{CAP-CoT} cycle).}
			(1) Domain: $\sqrt{x-1}$ requires $x \ge 1$, and equality to $x-3$ also requires $x-3 \ge 0$, so $x \ge 3$.
			(2) Square: $x-1 = (x-3)^2 \Rightarrow x^2 - 7x + 10 = 0 \Rightarrow x \in {2,5}$.
			(3) Constraint check: only $x=5$ satisfies $x \ge 3$; discard $x=2$.
			(4) Substitute $x=5$: $\sqrt{5-1}=2$ and $5-3=2$, so it holds.
			(5) \emph{Final answer: ${5}$.} \\
			\bottomrule
		\end{tabular}
		\label{tab:ex1}
	\end{table*}
	
	\begin{table*}[ht]
		\centering
		\small
		\setlength{\tabcolsep}{6pt}
		\renewcommand{\arraystretch}{1.1}
		\begin{tabular}{p{0.97\linewidth}}
			\toprule
			\textbf{Example B.2: Confusion-type hard negative for unit/quantity separation (GSM8K-style).} \\
			\midrule
			\midrule
			\textbf{Question.} Alice drives $100$ miles at $50$ mph and then $100$ miles at $25$ mph. What is her average speed for the whole trip? \\
			\textbf{Solver output $C_S^{(0)}$.}
			(1) Average speed is total distance divided by total time.
			(2) Time: $100/50 = 2$ hours, and $100/25 = 4$ hours.
			(3) Total distance: $200$ miles; total time: $6$ hours.
			(4) Average speed: $200/6 \approx 33.3$ mph. \\
			\textbf{Challenger output $C_C^{(0)}$ (Confusion error: average of rates).}
			(1) The trip has two speeds: $50$ mph and $25$ mph.
			(2) Take their average: $(50+25)/2 = 37.5$ mph.
			(3) \emph{Answer: $37.5$ mph.} \\
			\textbf{Feedback (step-aligned).}
			\textbf{For $C_S^{(0)}$:} The computation is correct, but the justification in Step (1) is brief. Add an explicit reminder that averaging rates requires weighting by time or distance; otherwise it becomes the confusion exploited by $C_C^{(0)}$.
			\textbf{For $C_C^{(0)}$:} The chain confuses “mean of two speeds” with “distance-weighted average speed,” ignoring that time differs across legs. \\
			\textbf{Solver output $C_S^{(1)}$ (after one \ac{CAP-CoT} cycle).}
			(1) Average speed is $\frac{\text{total distance}}{\text{total time}}$; we cannot average mph directly unless the legs have equal time (or we apply proper weighting).
			(2) Compute time per leg: $100/50=2$ hours and $100/25=4$ hours.
			(3) Total distance $=200$ miles, total time $=6$ hours.
			(4) Average speed $=200/6 \approx 33.3$ mph. \\
			\bottomrule
		\end{tabular}
		\label{tab:ex2}
	\end{table*}
	
	\begin{table*}[ht]
		\centering
		\small
		\setlength{\tabcolsep}{6pt}
		\renewcommand{\arraystretch}{1.1}
		\begin{tabular}{p{0.97\linewidth}}
			\toprule
			\textbf{Example B.3: Wrapper-type hard negative for guarding against hidden assumptions (BBH-style syllogism).} \\
			\midrule
			\midrule
			\textbf{Question.} Consider the argument: “All florps are glibs. Some glibs are drams. Therefore, some florps are drams.” Is the conclusion \emph{necessarily} true? \\
			\textbf{Solver output $C_S^{(0)}$.}
			(1) We know $\text{florps} \subseteq \text{glibs}$.
			(2) We also know $\exists x: x \in \text{glibs} \cap \text{drams}$.
			(3) \emph{Answer: Not necessarily true.} \\
			\textbf{Challenger output $C_C^{(0)}$ (Wrapper error: fluent but assumes overlap).}
			(1) Since all florps are glibs, florps live inside the set of glibs.
			(2) Some glibs are drams, so the “dram part” exists inside glibs.
			(3) Therefore florps must include at least one element in that dram part.
			(4) \emph{Answer: Yes, the conclusion must be true.} \\
			\textbf{Feedback (step-aligned).}
			\textbf{For $C_S^{(0)}$:} The final judgment is correct, but Step (3) lacks a concrete refutation. Add a counterexample to show why the conclusion does not follow (to prevent the wrapper-style assumption in $C_C^{(0)}$ from sounding persuasive).
			\textbf{For $C_C^{(0)}$:} The argument quietly assumes $\text{florps} \cap \text{drams} \neq \emptyset$ from $\text{florps} \subseteq \text{glibs}$ and $\text{glibs} \cap \text{drams} \neq \emptyset$, which is invalid without an additional premise. \\
			\textbf{Solver output $C_S^{(1)}$ (after one \ac{CAP-CoT} cycle).}
			(1) Premises: $\text{florps} \subseteq \text{glibs}$ and $\text{glibs} \cap \text{drams} \neq \emptyset$.
			(2) These do \emph{not} imply $\text{florps} \cap \text{drams} \neq \emptyset$.
			(3) Counterexample: let glibs $={a,b}$, florps $={a}$, drams $={b}$. Then all florps are glibs and some glibs are drams, but no florps are drams.
			(4) \emph{Final answer: the conclusion is not necessarily true.} \\
			\bottomrule
		\end{tabular}
		\label{tab:ex3}
	\end{table*}
	
	\section{Appendix B: Output Examples}
	\label{app2}
	In this section, we provide qualitative examples (Example B.1, B.2, B.3) with typical error types to demonstrate the optimization process of \ac{CAP-CoT}, showing a solver output, a challenger-generated hard negative, step-aligned feedback, and the improved solver output after one cycle. 

	\section{Appendix C: Error Strategy Coverage and Minimality Analysis}
	\label{app3}
	This appendix analyzes the error strategy design of the Adaptive Adversarial Challenger ($G_C$) in \ac{CAP-CoT} in detail, focusing on whether the cold-start error taxonomy is critical for the gains, whether \ac{CAP-CoT} depends on carefully engineered combinations of error types, and whether the feedback cycle can still start and evolve with only a minimal error signal.
	
	\subsection{Cold-start Error Taxonomy}
	As described in Section 3.2, the challenger is cold-started with a lightweight error taxonomy covering four common reasoning failure modes \{\emph{jump}, \emph{confusion}, \emph{fuzzy}, \emph{wrapper}\}. The purpose of this taxonomy is not to enumerate all possible errors, but to provide a minimal and operational bootstrap signal that makes the contrastive cycle start reliably. A key concern is whether \ac{CAP-CoT} relies on this multi-type taxonomy, or whether it mainly serves as a bootstrap mechanism. To test minimality, we run a single-error cold-start ablation. We use GPT-4o-mini as the backbone and evaluate on MATH, GSM8K, and BBH for three optimization cycles. The only change is that, in the first one to two cycles, the challenger is restricted to a single error type; all other settings (solver, feedback agent, decoding configuration) are kept the same. For inference, we use the final solver prompt with a single forward pass, without the challenger or feedback roles.
	
	Table~\ref{tab:single_error_coldstart} reports the final accuracies under different single-error cold starts. It can be observed that a single error type is sufficient to start \ac{CAP-CoT}, as each single-error setting reaches performance close to the full-taxonomy setting after two to three cycles across all three datasets. Also, differences among single-error settings are small (within about one point) and remain within the run-to-run standard deviation, suggesting that the method does not depend on any particular error type. The full taxonomy is slightly more stable on average, but with a limited margin, indicating that the primary role of the cold-start taxonomy is to make the early cycles more reliable, rather than to determine the final performance. These findings support the view that \ac{CAP-CoT} does not require a finely designed or comprehensive taxonomy. Instead, the cold-start strategy mainly triggers the adversarial feedback cycle, while later improvements are driven by feedback-conditioned strategy adaptation based on the solver's current behavior.
	
	\begin{table}[t]
		\caption{Single-error cold-start ablation on three datasets (mean$\pm$std over three runs). The challenger is restricted to one error type in the cold-start stage, while all other settings are identical.}
		\label{tab:single_error_coldstart}
		\centering
		\small
		\setlength{\tabcolsep}{3pt}
		\begin{tabular}{lccc}
			\toprule
			\textbf{Cold-start Error} & \textbf{MATH (\%)} & \textbf{GSM8K (\%)} & \textbf{BBH (\%)} \\
			\midrule
			\midrule
			Jump only      & 86.8$\pm$0.32 & 95.9$\pm$0.21 & 87.4$\pm$0.28 \\
			Confusion only & 86.5$\pm$0.35 & 96.0$\pm$0.18 & 87.2$\pm$0.31 \\
			Fuzzy only     & 86.6$\pm$0.37 & 95.7$\pm$0.24 & 87.3$\pm$0.34 \\
			Wrapper only   & 86.9$\pm$0.30 & 95.8$\pm$0.22 & 87.6$\pm$0.27 \\
			\midrule
			Full taxonomy & \textbf{87.2$\pm$0.29} & \textbf{96.1$\pm$0.17} & \textbf{87.9$\pm$0.25} \\
			\bottomrule
		\end{tabular}
	\end{table}

	\subsection{Strategy Evolution Beyond the Cold-start}
	
	Although the challenger starts with a fixed set of error types, its strategy is not constrained by that taxonomy in later cycles. In cycles two to three, we frequently observe derived error patterns that are not explicitly listed in the cold-start set, such as: omitting candidate validation (producing candidate answers but failing to check constraints), injecting implicit assumptions (using an unstated premise that is required for the conclusion), misinterpreting constraint direction (e.g., reversing an inequality or logical condition), and causal reversal (keeping local steps plausible while flipping causal direction). These patterns emerge as the feedback agent updates the challenger prompt to better match the solver's current blind spots, rather than being manually specified.
	
	\subsection{Case Studies}
	
	The qualitative examples in Appendix B illustrate this evolution. For MATH, errors can shift from a simple \textit{Jump} to a more specific failure, such as producing candidates without filtering them by domain constraints. For BBH-style logical tasks, errors can evolve from a generic \textit{Wrapper} into a concrete invalid inference, such as incorrectly concluding that an intersection is non-empty from a subset relation and a separate non-empty intersection. These cases align with the ablation results above, where \ac{CAP-CoT} can start from a minimal error signal, and then progressively discovers more fine-grained and more diagnostic failure modes through cycle-by-cycle feedback and challenger adaptation.
	
	
	\printcredits

	
	
	
	
\end{document}

%% file: CQILABInclusion/CQILAB-Notations.tex
\usepackage{xcolor}
\usepackage{braket}
\usepackage{bbm}
\usepackage{pifont} 
\usepackage{amssymb}

\DeclareMathAlphabet{\mathpzc}{OT1}{pzc}{m}{it}

\DeclareMathAlphabet{\mathitsf}{OML}{cmbr}{m}{it}
\DeclareMathAlphabet{\mathsf}{OT1}{cmbr}{m}{n}

\definecolor{darkgreen}{RGB}{0,128,0}

\makeatletter
\@ifundefined{IEEEQED}{}{%
	\AtBeginDocument{}%
}
\makeatother


%% file: CQILABInclusion/CQILAB-Acronyms.tex
\DeclareAcronym{1D}{
short = 1D,
long = one-dimensional
}

\DeclareAcronym{2D}{
short = 2D,
long = two-dimensional
}

\DeclareAcronym{3D}{
short = 3D,
long = three-dimensional
}

\DeclareAcronym{5G}{
short = 5G,
long = fifth generation
}

\DeclareAcronym{6G}{
short = 6G,
long = sixth-generation
}

\DeclareAcronym{AI}{
short = AI,
long = artificial intelligence
}

\DeclareAcronym{ANN}{
short = ANN,
long = artificial neural network
}

\DeclareAcronym{AR}{
short = AR,
long = augmented reality
}

\DeclareAcronym{AO}{
short = AO,
long = absorptive object
}

\DeclareAcronym{AWGN}{
short = AWGN,
long = additive white Gaussian noise
}

\DeclareAcronym{BB84}{
short = BB84,
long = Bennett and Brassard 1984
}

\DeclareAcronym{BEP}{
short = BEP,
long = bit error probability
}

\DeclareAcronym{BLIP}{
short = BLIP,
long = bootstrapping language-image pretraining
}

\DeclareAcronym{BPSK}{
short = BPSK,
long = binary phase-shift keying
}

\DeclareAcronym{BS}{
short = BS,
long = base station
}

\DeclareAcronym{BSM}{
short = BSM,
long = Bell-state measurement
}

\DeclareAcronym{CKA}{
short = CKA,
long = conference key agreement
}

\DeclareAcronym{CNN}{
short = CNN,
long = convolutional neural network
}

\DeclareAcronym{CNOT}{
short = CNOT,
long = controlled-NOT
}

\DeclareAcronym{CPS}{
short = CPS,
long = cyber-physical system
}

\DeclareAcronym{CPTP}{
short = CPTP,
long = completely positive trace-preserving
}

\DeclareAcronym{CPU}{
short = CPU,
long = central processing unit
}

\DeclareAcronym{CSI}{
short = CSI,
long = channel state information
}

\DeclareAcronym{CV}{
short = CV,
long = continuous-variable 
}

\DeclareAcronym{DNN}{
short = DNN,
long = deep neural network
}

\DeclareAcronym{DL}{
short = DL,
long = deep learning
}

\DeclareAcronym{DQL}{
short = DQL,
long = deep Q-learning
}

\DeclareAcronym{DRL}{
short = DRL,
alt = deep RL,
long = deep reinforcement learning
}

\DeclareAcronym{DT}{
short = DT,
long = digital twin
}

\DeclareAcronym{DV}{
short = DV,
long = discrete-variable 
}

\DeclareAcronym{FD}{
short = FD,
long = full-duplex
}

\DeclareAcronym{GAN}{
short = GAN,
long = generative adversarial network
}

\DeclareAcronym{GEO}{
short = GEO,
long = geostationary Earth orbit
}

\DeclareAcronym{GHZ}{
short = GHZ,
long = Greenberger--Horne--Zeilinger
}

\DeclareAcronym{GPS}{
short = GPS,
long = global positioning system
}

\DeclareAcronym{GPT}{
short = GPT,
long = generative pretrained transformer
}

\DeclareAcronym{GPU}{
short = GPU,
long = graphics processing unit 
}

\DeclareAcronym{GUS}{
short = GUS,
long = geometrically uniform symmetry
}

\DeclareAcronym{HQC}{
short = HQC,
long = hybrid quantum-classical
}

\DeclareAcronym{IIoT}{
short = IIoT,
alt = industrial IoT,
long = industrial Internet of Things
}

\DeclareAcronym{IoIV}{
short = IoIV,
long = Internet of intelligent Vehicles
}

\DeclareAcronym{IoT}{
short = IoT,
long = Internet of Things
}

\DeclareAcronym{ISAC}{
short = ISAC,
long = integrated sensing and communication
}

\DeclareAcronym{KPI}{
short = KPI,
long = key performance indicator
}

\DeclareAcronym{LEO}{
short = LEO,
long = low Earth orbit
}

\DeclareAcronym{LLaMA}{
short = LLaMA,
alt = large language model Meta AI,
long = LLM Meta AI
}

\DeclareAcronym{LLM}{
short = LLM,
long = large language model
}

\DeclareAcronym{LPIPS}{
short = LPIPS,
long = learned perceptual image patch similarity
}

\DeclareAcronym{LOCC}{
short = LOCC,
long = local operations and classical communication
}

\DeclareAcronym{LSTM}{
short = LSTM,
long = long short-term memory
}

\DeclareAcronym{MEO}{
short = MEO,
long = medium Earth orbit
}

\DeclareAcronym{MIMO}{
short = MIMO,
long = multiple-input multiple-output
}

\DeclareAcronym{mmWave}{
short = mmWave,
long = millimeter-wave
}

\DeclareAcronym{ML}{
short = ML,
long = machine learning
}

\DeclareAcronym{MLP}{
short = MLP,
long = multi-layer perceptron
}

\DeclareAcronym{MPA}{
short = MPA,
long = mean pixel accuracy
}

\DeclareAcronym{MR}{
short = MR,
long = mixed reality
}

\DeclareAcronym{MSE}{
short = MSE,
long = mean squared error
}

\DeclareAcronym{NISQ}{
short = NISQ,
long = noisy intermediate-scale quantum
}

\DeclareAcronym{NMSE}{
short = NMSE,
alt = normalized MSE,
long = normalized mean squared error
}

\DeclareAcronym{NLP}{
short = NLP,
long = natural language processing
}

\DeclareAcronym{NTN}{
short = NTN,
long = non-terrestrial network
}

\DeclareAcronym{PISQ}{
short = PISQ,
long = perfect intermediate-scale quantum
}

\DeclareAcronym{POVM}{
short = POVM,
long = positive operator-valued measure
}

\DeclareAcronym{PQC}{
short = PQC,
long = parametrized quantum circuit
}

\DeclareAcronym{QAA}{
short = QAA,
long = quantum anonymous authentication
}

\DeclareAcronym{QAB}{
short = QAB,
long = quantum anonymous broadcast
}

\DeclareAcronym{QAC}{
short = QAC,
long = quantum anonymous communication
}

\DeclareAcronym{QACKA}{
short = QA-CKA,
alt = quantum anonymous CKA,
long = quantum anonymous conference key agreement
}

\DeclareAcronym{QACD}{
short = QACD,
long = quantum anonymous collision detection
}

\DeclareAcronym{QAE}{
short = QAE,
long = quantum anonymous entanglement
}

\DeclareAcronym{QAIA}{
short = QAIA,
long = quantum anonymous identity authentication
}

\DeclareAcronym{QAIR}{
short = QAIR,
long = quantum anonymous information retrieval
}

\DeclareAcronym{QAM}{
short = QAM,
long = quadrature amplitude modulation
}

\DeclareAcronym{QAN}{
short = QAN,
long = quantum anonymous network
}

\DeclareAcronym{qAN}{
short = QAN,
long = quantum anonymous notification
}

\DeclareAcronym{QANO}{
short = QANO,
long = quantum anonymous notification
}

\DeclareAcronym{QAOA}{
short = QAOA,
long = quantum approximate optimization algorithm
}

\DeclareAcronym{QAP}{
short = QAP,
long = quantum anonymous publication
}

\DeclareAcronym{QAR}{
short = QAR,
long = quantum anonymous ranking
}

\DeclareAcronym{QAS}{
short = QAS,
long = quantum anonymous sensing
}

\DeclareAcronym{QAT}{
short = QAT,
long = quantum anonymous teleportation
}

\DeclareAcronym{QAV}{
short = QAV,
long = quantum anonymous voting
}

\DeclareAcronym{QCC}{
short = QCC,
long = quantum covert communication
}

\DeclareAcronym{QCRB}{
short = QCRB,
long = quantum Cram\'er--Rao bound
}

\DeclareAcronym{QCS}{
short = QCS,
long = quantum clock synchronization
}

\DeclareAcronym{QEC}{
short = QEC,
long = quantum error correction
}

\DeclareAcronym{QED}{
short = QED,
long = quantum entanglement distribution
}

\DeclareAcronym{QFI}{
short = QFI,
long = quantum Fisher information
}

\DeclareAcronym{QFIM}{
short = QFIM,
alt = QFI matrix,
long = quantum Fisher information matrix
}

\DeclareAcronym{QFT}{
short = QFT,
long = quantum Fourier transform
}

\DeclareAcronym{QIoT}{
short = QIoT,
alt = quantum IoT,
long = quantum Internet of Things
}

\DeclareAcronym{QKD}{
short = QKD,
long = quantum key distribution
}

\DeclareAcronym{QLSTM}{
short = QLSTM,
alt = quantum LSTM,
long = quantum long short-term memory
}

\DeclareAcronym{QML}{
short = QML,
alt = quantum ML,
long = quantum machine learning
}

\DeclareAcronym{QNN}{
short = QNN,
long = quantum neural network
}

\DeclareAcronym{QOC}{
short = QOC,
long = quantum optimal control
}

\DeclareAcronym{QPU}{
short = QPU,
long = quantum processing unit
}

\DeclareAcronym{QSC}{
short = QSC,
alt = quantum SC,
long = quantum semantic communication
}

\DeclareAcronym{QSN}{
short = QSN,
long = quantum sensing network
}

\DeclareAcronym{QUBO}{
short = QUBO,
long = quadratic unconstrained binary optimization
}

\DeclareAcronym{ReLU}{
short = ReLU,
long = rectified linear unit
}

\DeclareAcronym{ResNet}{
short = ResNet,
long = residual network
}

\DeclareAcronym{RGB}{
short = RGB,
long = red-green-blue
}

\DeclareAcronym{RL}{
short = RL,
long = reinforcement learning
}

\DeclareAcronym{RMSE}{
short = RMSE,
alt = root MSE,
long = root mean squared error
}

\DeclareAcronym{RNN}{
short = RNN,
long = recurrent neural network
}

\DeclareAcronym{RSA}{
short = RSA,
long = Rivest--Shamir--Adleman
}

\DeclareAcronym{SAM}{
short = SAM,
long = segment anything model
}

\DeclareAcronym{SC}{
short = SC,
long = semantic communication
}

\DeclareAcronym{SEP}{
short = SEP,
long = symbol error probability
}

\DeclareAcronym{SNR}{
short = SNR,
long = signal-to-noise ratio
}

\DeclareAcronym{SRM}{
short = SRM,
long = square-root measurement
}

\DeclareAcronym{SQL}{
short = SQL,
long = standard quantum limit
}

\DeclareAcronym{SVM}{
short = SVM,
long = support vector machine
}

\DeclareAcronym{SwinT}{
short = SwinT,
long = shifted window transformer
}

\DeclareAcronym{THz}{
short = THz,
long = terahertz
}

\DeclareAcronym{TN}{
short = TN,
long = terrestrial network
}

\DeclareAcronym{UAV}{
short = UAV,
long = unmanned aerial vehicle
}

\DeclareAcronym{U-Net}{
short = U-Net,
long = U-shaped network
}

\DeclareAcronym{VAE}{
short = VAE,
long = variational autoencoder
}

\DeclareAcronym{V2I}{
short = V2I,
long = vehicle-to-infrastructure
}

\DeclareAcronym{V2X}{
short = V2X,
long = vehicle-to-everything
}

\DeclareAcronym{VFM}{
short = VFM,
long = vision foundation model
}

\DeclareAcronym{ViT}{
short = ViT,
long = vision transformer
}

\DeclareAcronym{VQA}{
short = VQA,
long = variational quantum algorithm
}

\DeclareAcronym{VQC}{
short = VQC,
long = variational quantum circuit
}

\DeclareAcronym{VQS}{
short = VQS,
long = variational quantum sensing
}

\DeclareAcronym{VR}{
short = VR,
long = virtual reality
}

\DeclareAcronym{YOLO}{
short = YOLO,
long = you only look once
}


%% file: LatexInclusion/CAP-CoT-Acronyms.tex
\DeclareAcronym{CoT}{
	short = CoT,
	long  = chain-of-thought
}

\DeclareAcronym{QA}{
	short = QA,
	long  = question answering
}

\DeclareAcronym{SFPR}{
	short = SFPR,
	long  = structured feedback prompt refinement
}

\DeclareAcronym{CoT-SC}{
	short = CoT-SC,
	long  = self-consistency chain-of-thought
}

\DeclareAcronym{CCoT}{
	short = CCoT,
	long  = contrastive chain-of-thought
}

\DeclareAcronym{ToT}{
	short = ToT,
	long  = tree of thoughts
}

\DeclareAcronym{GoT}{
	short = GoT,
	long  = graph of thoughts
}

\DeclareAcronym{FoT}{
	short = FoT,
	long  = Forest-of-Thought
}

\DeclareAcronym{AoT}{
	short = AoT,
	long  = atom of thoughts
}

\DeclareAcronym{CAP-CoT}{
	short = CAP-CoT,
	long  = cycle adversarial prompt optimization
}
\DeclareAcronym{AP}{
	short = AP,
	long  = analogical prompting
}

\DeclareAcronym{AFlow}{
	short = AFlow,
	long  = analogical flow
}

\DeclareAcronym{PAgent}{
	short = PAgent,
	long  = PromptAgent
}

%% file: LatexInclusion/CAP-CoT-Notations.tex
\usepackage{xcolor}
\usepackage{braket}
\usepackage{bbm}
\usepackage{pifont}